\documentclass{article}

\usepackage{arxiv}

\usepackage[utf8]{inputenc} 
\usepackage[T1]{fontenc}    
\usepackage{hyperref}       
\usepackage{url}            
\usepackage{booktabs}       
\usepackage{amsfonts}       
\usepackage{nicefrac}       
\usepackage{microtype}      
\usepackage{lipsum}		
\usepackage{graphicx}
\usepackage{natbib}
\usepackage{doi}

\usepackage{amsmath}
\usepackage{amssymb}
\usepackage{algorithm}
\usepackage{algpseudocode}
\usepackage{subcaption}
\usepackage{graphicx}
\usepackage{array}
\usepackage{tabularx}
\usepackage{booktabs}

\usepackage{multirow}

\title{IPG: Incremental Patch Generation for Generalized Adversarial Patch Training}

\date{} 					

\author{ \href{https://orcid.org/0009-0005-7473-743X}{\includegraphics[scale=0.06]{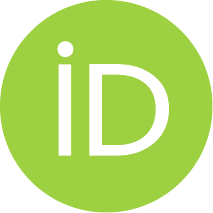}\hspace{1mm}Wonho Lee}\thanks{This work was supported by the Institute of Information \& Communications Technology Planning \& Evaluation (IITP) grant funded by the Korea government (MSIT) (No. 2021-0-00511, Robust AI and Distributed Attack Detection for Edge AI Security) and (No. RS-2024-00398353, Development of Countermeasure Technologies for Generative AI Security Threats).} \\
	Department of Software\\
	Soongsil University\\
	\texttt{hoho0907@soongsil.ac.kr} \\
	\And
	\href{https://orcid.org/0000-0002-7123-4467}{\includegraphics[scale=0.06]{orcid.pdf}\hspace{1mm}Hyunsik Na} \\
	Department of Software\\
	Soongsil University\\
	\texttt{rnrud7932@soongsil.ac.kr} \\
        \And
        \href{https://orcid.org/0009-0009-8932-6333}{\includegraphics[scale=0.06]{orcid.pdf}\hspace{1mm}Jisu Lee} \\
	Department of Software\\
	Soongsil University\\
	\texttt{connandgo@soongsil.ac.kr} \\
        \And
        \href{https://orcid.org/0000-0002-1438-0265}{\includegraphics[scale=0.06]{orcid.pdf}\hspace{1mm}Daeseon Choi}\thanks{Corresponding Author} \\
	Department of Software\\
	Soongsil University\\
	\texttt{sunchoi@ssu.ac.kr} \\
}

\date{}

\renewcommand{\headeright}{}
\renewcommand{\undertitle}{}
\renewcommand{\shorttitle}{IPG: Incremental Patch Generation for Generalized Adversarial Patch Training}

\begin{document}
\maketitle

\begin{abstract}
The advent of adversarial patches poses a significant challenge to the robustness of AI models, particularly in the domain of computer vision tasks such as object detection. In contradistinction to traditional adversarial examples, these patches target specific regions of an image, resulting in the malfunction of AI models. This paper proposes Incremental Patch Generation (IPG), a method that generates adversarial patches up to 11.1 times more efficiently than existing approaches while maintaining comparable attack performance. The efficacy of IPG is demonstrated by experiments and ablation studies including YOLO's feature distribution visualization and adversarial training results, which show that it produces well-generalized patches that effectively cover a broader range of model vulnerabilities. Furthermore, IPG-generated datasets can serve as a robust knowledge foundation for constructing a robust model, enabling structured representation, advanced reasoning, and proactive defenses in AI security ecosystems.  The findings of this study suggest that IPG has considerable potential for future utilization not only in adversarial patch defense but also in real-world applications such as autonomous vehicles, security systems, and medical imaging, where AI models must remain resilient to adversarial attacks in dynamic and high-stakes environments.
\end{abstract}


\section{Introduction}
Since the advent of Convolutional Neural Networks \cite{CNN_krizhevsky2012}, Artificial Intelligence(AI) has become the dominant technology in the field of computer vision. AI has achieved state-of-the-art performance in various tasks such as image classification, object detection, and segmentation, and is widely applied in real-world scenarios. In particular, for real-time object detection tasks, the You Only Look Once (YOLO) series \cite{yolo_redmon2016} has been effectively utilized in autonomous driving and unmanned retail environments due to its fast inference speed and high accuracy.

However, alongside advancements in AI, security vulnerabilities have also become a growing concern. Among these, adversarial patches \cite{adversarialpatch_brown2017} are regarded as one of the most critical threats, as they can physically manifest in real-world environments and cause severe model misbehavior. Adversarial patches can lead to catastrophic outcomes by inducing malfunctions in AI models.

To address this issue, various defense strategies against adversarial patch attacks have been proposed \cite{certified_chiang2020,randomizedsmotthing_levine2020,watermarking_hayes2018,LGS_naseer2019,VaN_gittings2020,locationoptimaadvtrain_rao2020}. These can be broadly categorized into three main groups: (1) Certified Defense, (2) Detection and Removal, and (3) Robust Training. Among these strategies, (1) and (2) necessitate additional latency and computational overhead in real-time object detection systems. Consequently, Robust Training, which aims to enhance the model's resilience during training, is the most efficient approach. In the context of robust training, adversarial training represents a prominent method for defending against adversarial patches. This approach entails integrating adversarial patch-applied data into the model’s training process, thereby enhancing its robustness against such attacks. Effective adversarial training for adversarial patches necessitates a substantial number of patches generalized by various factors, including the location, angle, and brightness of the patch. Furthermore, it is imperative to comprehensively investigate the model's vulnerable space, wherein adversarial patches can be optimally positioned.

The generation of adversarial patches entails a multitude of considerations beyond those associated with traditional adversarial examples. These include factors such as patch angle, position, size, and the area of the object within the image. Generally, adversarial patches are optimized using the attacker's entire dataset, producing a single patch. However, current methods converge to similar areas due to optimization-based objective functions, limiting their capacity to address heterogeneous vulnerabilities in models, thus reducing their generalization. Consequently, models trained with such patches may have blind spots based on parameters like patch position and angle. Although creating a diverse set of adversarial patches could mitigate these issues, existing methods are inefficient, requiring substantial time for each patch.

To address these issues, we propose an efficient and generalized patch generation method for adversarial training named Incremental Patch Generation (IPG). To the best of our knowledge, no optimized method exists for efficiently generating diverse adversarial patches. IPG can generate patches up to 11.1 times faster than existing methods while maintaining comparable attack performance. We compare the distributions of traditional adversarial patches and IPG patches using Principal Component Analysis (PCA) and t-distributed Stochastic Neighbor Embedding (t-SNE) techniques and find that IPG generates more generalized patches across multiple factors. This generalization ability allows IPG to improve the inherent robustness of the model in adversarial training. The results of adversarial training experiments using IPG show that IPG can cover a wider spectrum of security vulnerabilities in adversarial patches, which significantly improves the inherent robustness of the model, making it more resilient in real-world applications. We show that IPG is a suitable method for generalized adversarial training.

Furthermore, the generalized adversarial patch dataset generated by IPG serves as a robust knowledge foundation that can be integrated into a robust framework. By organizing adversarial patch-related knowledge systematically, IPG enables the construction of a resilient knowledge structure, enhancing interoperability and knowledge representation. This facilitates advanced reasoning and inference, ultimately promoting proactive defense mechanisms and informed decision-making within AI security ecosystems. Our adversarial training experiments confirm that IPG covers a broader spectrum of vulnerabilities, significantly improving model robustness, thereby demonstrating IPG's effectiveness as a foundational component for AI security.
The contributions of this study are as follows:

\begin{itemize}
    \item To the best of our knowledge, we present the first efficient and effective optimization method, IPG, for constructing samples of adversarial patch training for robust training.

    \item We demonstrate that the IPG method covers a significantly broader range of vulnerability space than existing patch generation method. This superior generalization performance is shown through detailed analysis and visualization.

    \item We conducted adversarial training with patches generated by IPG, demonstrating that IPG can enhance the inherent robustness of AI models.

    \item The findings of ablation studies on IPG have provided insights and potential directions for future research on generalized adversarial patch training.

    \item IPG-generated datasets facilitate the development of robust knowledge, supporting enhanced semantic interoperability and proactive AI security measures.

\end{itemize}

\section{Related Work}
\subsection{Adversarial Patch Attack}
Adversarial attacks represent a prominent method of targeting deep learning models. This is achieved by injecting minimal perturbations into input data in a way that is imperceptible to humans, causing the model to malfunction. Among these, adversarial patches, initially proposed by Brown et al. \cite{adversarialpatch_brown2017}, represent a robust form of adversarial attack that can withstand physical-world transformations or distortions with small image patches. In contrast to traditional adversarial examples, which inject perturbations across the entire image, adversarial patches facilitate more straightforward attacks by attaching a patch to an arbitrary location within the input image.

Since the introduction of adversarial patches, various studies have concentrated on enhancing their capacity to attack while simultaneously reducing their visibility by the human eye. For example, Karmon et al. \cite{lavan_karmon2018} proposed LaVAN(Localized and Visible Adversarial Noise), which allows for the creation of smaller adversarial patches. Similarly, Chindaudom et al. \cite{qrpatch_chindaudom2020} introduced QRPatch, designed in the shape of a QR code to make it more difficult for humans to detect.

The initial applications of adversarial patches were primarily concerned with image classification. In a pioneering contribution, Liu et al. proposed DPatch, an adversarial patch attack against object detection models \cite{daptch_liu2018}. To successfully attack object detection models, it is necessary to disrupt both the model’s bounding box regression and object classification. Adversarial patches are particularly dangerous in real-time object detection tasks, as they involve manipulating the image in a way that causes the model to perceive the patch as part of the object. For instance, in practical scenarios such as fully unmanned stores, adversarial patches could be employed to hide objects, resulting in missed detections or misclassifications and consequently leading to considerable economic damage.

However, when generating adversarial patches, attackers seek to generate an optimal patch by leveraging as many exploitable factors as possible. Consequently, optimization-based objective functions are frequently employed, resulting in the generation of patches that converge to similar regions. This limitation in generalization and introduction of bias are inherent to this approach. Moreover, although utilizing the entire dataset may enhance the attack success rate, generating a single patch necessitates a considerable investment of time, which represents a substantial limitation.

\subsection{Adversarial Patch Defense}
As AI models become widely deployed in security systems and safety-critical domains, adversarial patches are demonstrated to pose a significant and tangible threat in real-world scenarios, the development of defenses against adversarial patches has become a crucial and urgent priority. Defenses against adversarial patches can be broadly categorized into three main groups: Certified Defense, Detection and Removal, and Robust Training.

Various proposed defenses in the context of adversarial patches have been based on input pre-processing during inference time. Chiang et al. \cite{certified_chiang2020} argued that such defense techniques are vulnerable to white-box attacks, thereby encouraging the development of Certified Defenses. Certified Defense methods provide a mathematical guarantee that the model will provide a certain level of protection against a given attack. This guarantee extends not only to patch attacks but also to a range of potential threats. However, Certified Defense methods are typically computationally expensive, as they often entail evaluating extreme bounds to certify the model under worst-case scenarios. For example, Levine and Feizi's randomized smoothing \cite{randomizedsmotthing_levine2020} added random noise to the input data and averaged the predictions over multiple runs to achieve robust results. This pure defense strategy did not rely on patch location information and provided significant robustness against generic patch attacks.

Adversarial patches typically increase high-frequency perturbations in a localized region of the image, and they typically do not overlap with salient objects \cite{APSurvey_sharma2022}. For this reason, when a patch attack occurs, a dense clustering of perturbations in the area surrounding the patch can be observed in the saliency map. This allows for the detection and removal of such attacks using saliency maps. Hayes \cite{watermarking_hayes2018} proposed a defense strategy inspired by digital watermarking removal processes, employing both non-blind and blind image inpainting techniques to defend against adversarial patches. This method leveraged the saliency map of the image to remove small patches and mask adversarial perturbations. Similarly, Naseer et al. \cite{LGS_naseer2019} proposed Local Gradient Smoothing(LGS) to suppress highly activated and perturbed regions in the image without affecting salient objects. LGS processed the entire image globally while focusing on local regions, achieving robustness with minimal damage to clean accuracy.

In contrast to the previous two methods, adversarial training in robust training serves to enhance the model's robustness against adversarial examples through the incorporation of supplementary training. Gittings et al. \cite{VaN_gittings2020} were the first to propose Vax-a-Net(VaN), a method capable of defending against patch attacks during the training process. VaN employs a Deep Convolutional Generative Adversarial Network(DC-GAN) \cite{DCGAN_radford2015} to generate effective adversarial patches, which are then used to simultaneously train the model to defend against these patches, thereby providing a defense for image classification tasks.

However, the current state of adversarial training solutions presents significant limitations. Adversarial patch attacks tend to occur exclusively at specific locations and in specific shapes, increasing the probability that the model will exhibit robustness only to patches observed during training. Consequently, the model may remain vulnerable to unknown patch shapes or configurations not encountered during training. In particular, maintaining robustness against adversarial patches is a challenging task due to the variability in patch placement. To address this challenge, Rao et al. \cite{locationoptimaadvtrain_rao2020} proposed a method that explicitly optimizes the location of patches within the image to enhance the attack's effectiveness. This approach was subsequently applied in adversarial training to reduce the vulnerable regions susceptible to adversarial patches. Nevertheless, this approach was initially applied to image classification tasks. In the context of object detection tasks, the vast number of potential patch placement combinations, contingent on the object's position within the image, makes it impractical to generate a new patch for each iteration. This represents a limitation in the applicability of this method to object detection. Moreover, the training process necessitates the implementation of patches of varying sizes and shapes, which considerably prolongs both the time required for patch generation and the training time needed to learn from these patches. Consequently, this results in elevated computational costs and diminished efficiency. Therefore, the achievement of comprehensive defense against patch attacks remains an elusive goal \cite{APSurvey_sharma2022}.

\section{Method}
\subsection{Defender Knowledge}
We assumed that the defender is the service provider who builds and trains the model, thereby having access to the model's structure, parameters, and other internal components, which collectively grant them white-box privileges. It is generally accepted that white-box attacks are more powerful than black-box attacks, yet the defender is unaware of the attacker's level of knowledge. In this study, we assume that the attacker has full knowledge of the target model's parameters, architecture, and weights and can perform a white-box attack. This assumption allows us to investigate the most difficult attack scenarios, where the attacker possesses comprehensive knowledge of the model, which presents the greatest challenge for defense strategies. 

\subsection{Incremental Patch Generation}
\begin{figure}[t]
\centering
 \centering
    \begin{minipage}{0.4\textwidth}
        \centering
        \includegraphics[scale=0.3]{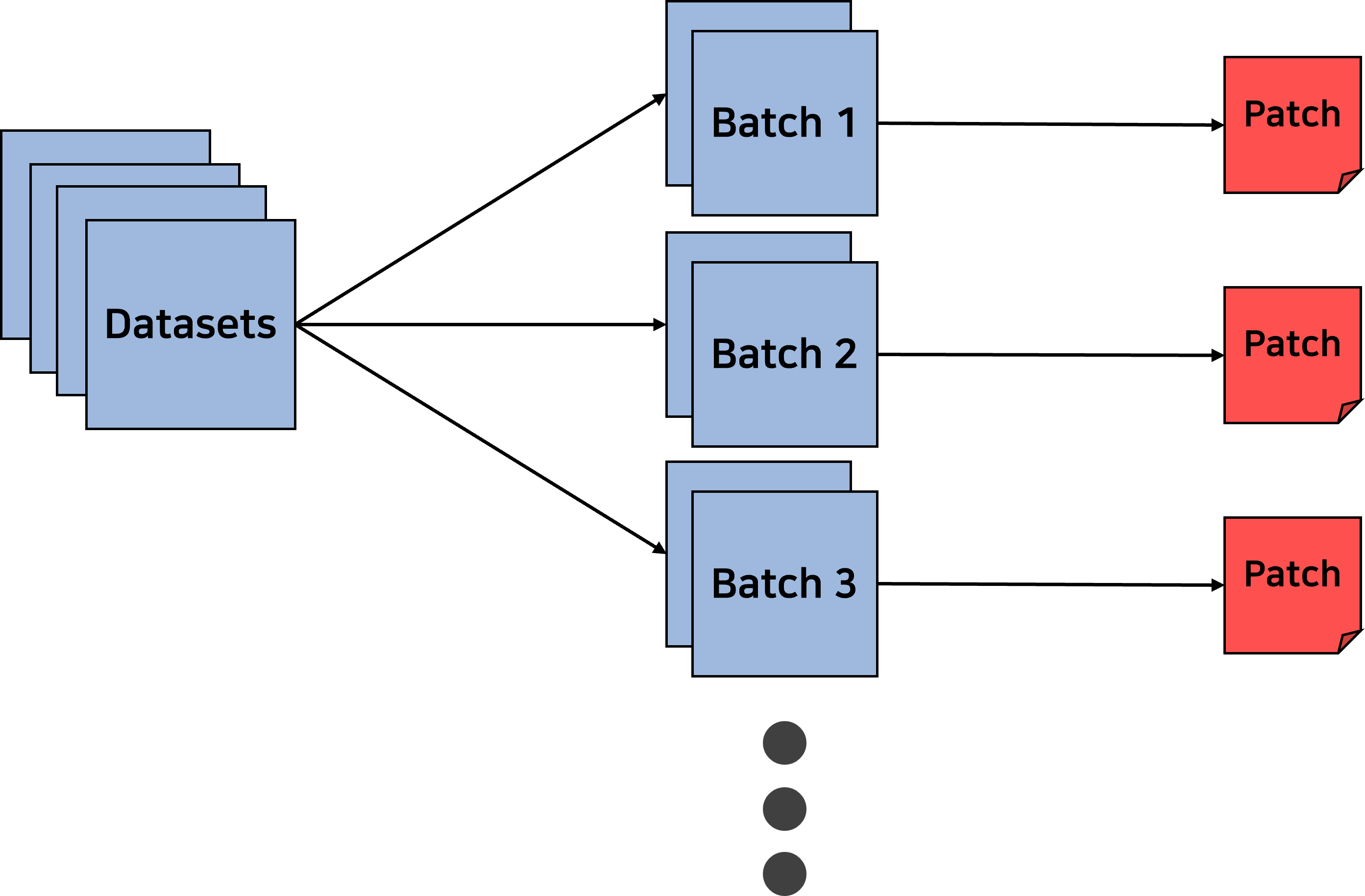} \\
        (\textbf{a})
    \end{minipage}%
    \hfill
    \begin{minipage}{0.5\textwidth}
        \centering
        \includegraphics[scale=0.3]{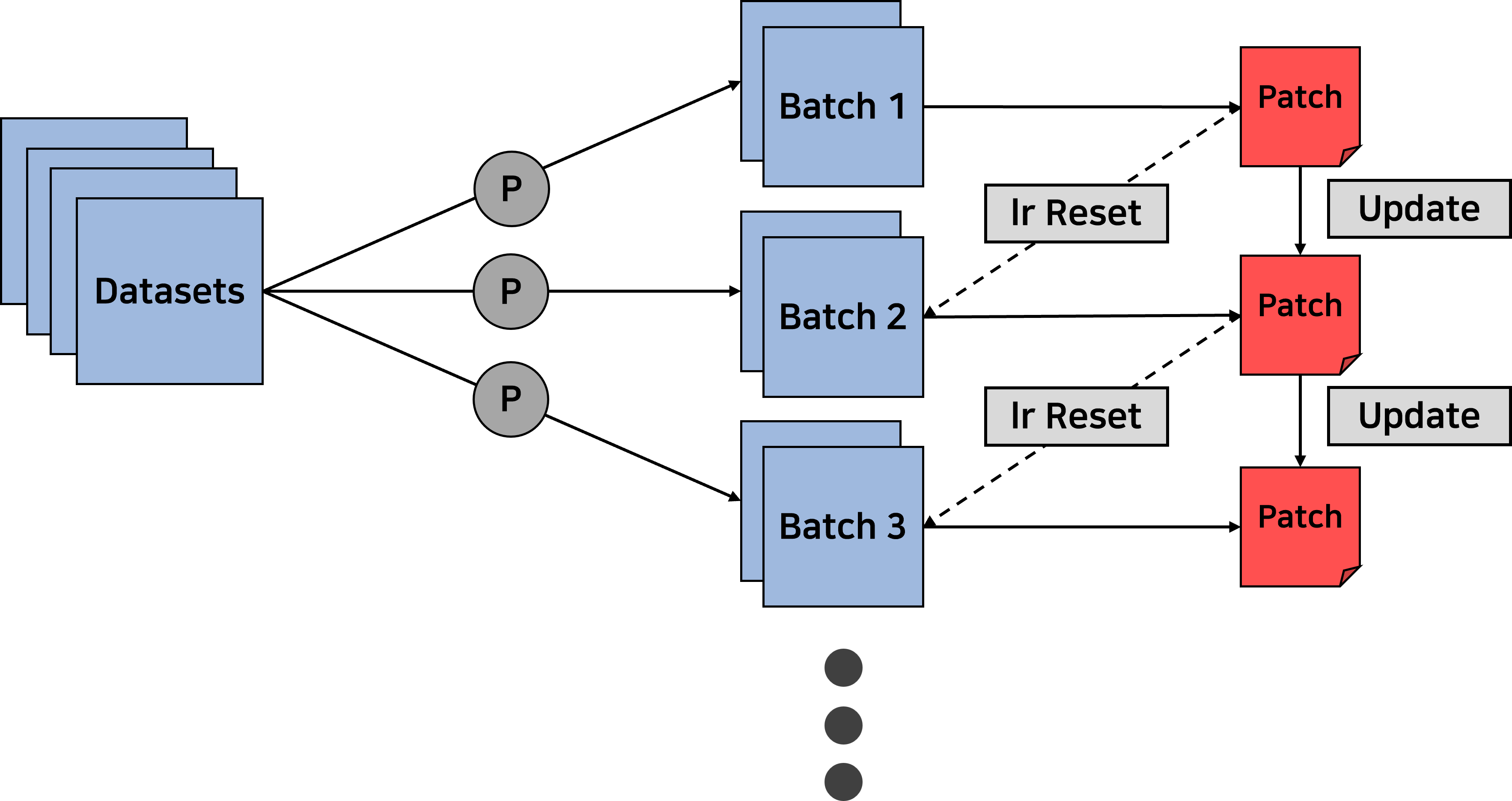} \\
        (\textbf{b})
    \end{minipage}%
\caption{The procedure of patch generation (\textbf{a}) Existing adversarial patch generation procedure (\textbf{b}) IPG procedure. P are the poisson sampler.}
\label{fig:procedure}
\end{figure}

Existing adversarial patches are generated as shown in Fig. \ref{fig:procedure}a.
The generation of the patch is accomplished by utilizing the entirety of the accessible dataset within a single epoch. The generation of a single adversarial patch with a high attack success rate necessitates a considerable investment of time, as the loss function must converge. Furthermore, the adversarial patch generated through this process becomes dependent on the images utilized during its creation and tends to exhibit a bias towards specific vulnerabilities inherent to the model.

In this paper, we propose Incremental Patch Generation (IPG), as shown in Fig. \ref{fig:procedure}b., to address the shortcomings of existing methodologies. IPG generates patches by extracting a subset of the available data. During the extraction process, the use of a random sampler may result in dependency on the data utilized in that particular batch. To address this issue, we employ a Poisson Sampler which has been shown to enhance generalization performance. The algorithm for the Poisson sampler is presented in Figure 2, which is inspired by the approaches proposed by Abadi et al. \cite{poisson_abadi2016} and Chua et al. \cite{poisson_chua2024}. The Poisson sampler selects data from the entire image set based on a certain probability, thereby ensuring that the amount of data extracted at each step varies slightly. This allows for the generation of adversarial patches that are not batch dependent.

The extracted data is then employed in the generation of patches through a process of 200 epochs per batch, with the learning rate decreasing in a progressive manner. Upon completion of patch generation for a given batch, a new batch is selected using the Poisson sampler, the learning rate is reset, and the patch is constantly updated. This process produces adversarial patches that are not dependent on the data used in specific batches, thereby enhancing generalization performance. Additionally, since fewer samples are utilized during the generation process, IPG is more efficient, enabling the creation of a larger number of patches in a shorter amount of time.

\algtext*{EndFor}
\begin{algorithm}[t]
\caption{An algorithm of Poisson Sampler $\mathcal{P}_{d,T}$}
\label{alg:Poisson sampler}
\textbf{Parameters:} Sample size $d$, number of patches to generate $T$. \\
\textbf{Input:} Number of all datapoints $n$. \\
\textbf{Output:} Seq. of batches $S_1, \dots, S_T \subseteq [n]$.
\begin{algorithmic}
    \For{$t = 1, \dots, T$}
        \State $S_t \gets \emptyset$
        \For{$i = 1, \dots, n$}
            \State $S_t \gets \begin{cases} S_t \cup \{i\} & \text{with probability } d/n \\ S_t & \text{with probability } 1 - d/n \end{cases}$
        \EndFor
    \EndFor
    \State \textbf{return} $S_1, \dots, S_T$
\end{algorithmic}
\end{algorithm}

\subsection{Attack Objective Function}
In the process of designing adversarial patches for object detection models such as the target model Yolov5l6, the attacker first establishes the attack goal and then defines the attack objective function in accordance with this goal. In this paper, we specifically executed a hiding attack, designed to evade detection by attaching adversarial patches that prevent objects from being recognized, on object detection models. This attack represents one of the various attacks that can target object detection models. The loss function for the hiding attack is defined as Equation 1 in order to minimize the object's confidence score ($x_cls$) and objectiveness score ($x_obj$) \cite{loss1_hu2021,loss2_shrestha2023}.

\begin{equation}
\tag{1}
\arg\min_p\mathbb{E}_{x \in X, l \in L, t \in T} \left( \max({x_{cls}\times x_{obj}})\right)
\end{equation}

Here, the position of each image, denoted by $l$, and the random transformation, represented by $t$, are used to update each pixel of the patch, which is represented by the variable $p$. And $\mathbb{E}$ denotes the expected value of the objective function.

\section{Experiment}
\subsection{Target Model}
Object detection models are among the most prevalent applications of AI in real-world scenarios. Adversarial patches pose a significant threat in physical environments, especially in the context of real-time object detection. Accordingly, the Yolov5l6 model \cite{yolov5}, which is capable of live object detection, was selected as the target model, and the MS COCO 2017 dataset \cite{coco_lin2014} was employed as the training set.

Additionally, to ascertain the robustness of IPG in relation to data-augmented models, the Yolov5l6 model was trained on the same dataset with the incorporation of data augmentation. To enhance the robustness of the model against common occlusions, we employed data augmentation, including multiplication, affine transformation, and dropout with Gaussian noise substitution. The performance of the target models is shown in Table \ref{table:yolov5l6_performance}.

\begin{table}[t] 
\caption{Performance of Yolov5l6 Target Model}
\vspace{3pt}
\label{table:yolov5l6_performance}
\newcolumntype{C}{>{\centering\arraybackslash}X}
\begin{tabularx}{\textwidth}{CCCCCC}
\toprule
\textbf{Dataset}&\textbf{mAP50}&\textbf{mAP50-95}&\textbf{Precision}&\textbf{Recall}&\textbf{F1-Score}\\
\midrule
COCO\_origin      & 0.658 & 0.470    &  0.744    & 0.596  & 0.662   \\
COCO\_aug         & 0.603 & 0.465    &  0.728    & 0.603  & 0.660   \\
\bottomrule
\end{tabularx}
\end{table}

\subsection{Datasets}
To generate adversarial patches, it is necessary to have access to data that adheres to the same distribution as that used for the model's training data. While access to the complete training dataset would facilitate the development of more effective patches, this is frequently unfeasible. Therefore, in this study, we assume that only 20\% of the total training data is accessible. Accordingly, 23,453 images were randomly extracted from the MS COCO 2017 dataset \cite{coco_lin2014}, which corresponds to 20\% of the training data, for the generation of patches.

\subsection{Implementation Details}
All experiments were conducted on an NVIDIA GeForce RTX 4090 GPU with the Python 3.12.4, PyTorch 2.3.1, and Torchvision 0.18.1 software environments. The adversarial patches were generated with a size of (64, 64), and the initial patches were initialized with random noise. To optimize the attack objective function, the Adam optimizer and a StepLR scheduler were employed. The learning rate was reduced from an initial value of 0.2 to 0.001 by the 200th epoch. To assess the efficacy of the training process, the MS COCO 2017 \cite{coco_lin2014} validation set was employed, with the proportion of the image occupied by the adversarial patches was probabilistically fixed within a range of 0.25 to 0.30.

\subsection{Evaluation Metrics \& Visualization}
To assess the efficacy of the generated adversarial patches, two metrics and one visualization technique were employed.
First, in order to assess the effectiveness of the adversarial patches as a means of attack, the Attack Success Rate (ASR) was calculated. The ASR was calculated using the MS COCO 2017 \cite{coco_lin2014} test set. In a hiding attack, the ASR can be defined as follows: $ASR = 1 - (TPR/N)$, where $N$ is the number of objects in the test image and $TPR$ is the True-Positive Rate(TPR) of the model. As IPG generates a substantial number of patches simultaneously, we assessed the mean ASR of all generated adversarial patches.

Secondly, the efficiency of the adversarial patch generation process was evaluated by measuring the patch generation rate relative to the time taken. The efficiency of the process was defined as the ratio of the number of generated patches to the time spent generating them, expressed as \textit{Efficiency = PatchAmount/GenHour} where $PatchAmount$ is the number of patches generated during that time, $GenHour$ is the time spent generating patches.

Lastly, to analyze the generalization of the generated adversarial patches, we employed Principal Component Analysis(PCA) \cite{PCA_mackiewicz1993} and t-distributed Stochastic Neighbor Embedding(t-SNE) \cite{tsne_hinton2002} for visualization purposes.
In order for adversarial patches to have an attack effect, they must be injected with the input of the model.
Consequently, the adversarial patches to be analyzed were attached to the same location in identical images and then input into the target model to extract features from backbone layers. We applied PCA to the extracted features, followed by two-dimensional T-SNE visualization, which enabled us to compare the generalization performance of the adversarial patches.

\begin{table}[t] 
\caption{Performance of Origin Patch and IPG Patch}
\vspace{3pt}
\label{table:result_IPG}
\newcolumntype{C}{>{\centering\arraybackslash}X}
\begin{tabularx}{\textwidth}{CCCCC}
\toprule
\textbf{Patch Method}&\textbf{Gen Time(hour)}&\textbf{Patch Amount}&\textbf{Efficiency}&\textbf{ASR}\\
\midrule
Origin         &  49.40         & 1            &  0.020     & 0.669 \\
IPG(ours)      & 112.55         & 25           &  0.222     & 0.600 \\
\bottomrule
\end{tabularx}
\end{table}
\begin{figure}[t]
\centering
\includegraphics[scale=0.50]{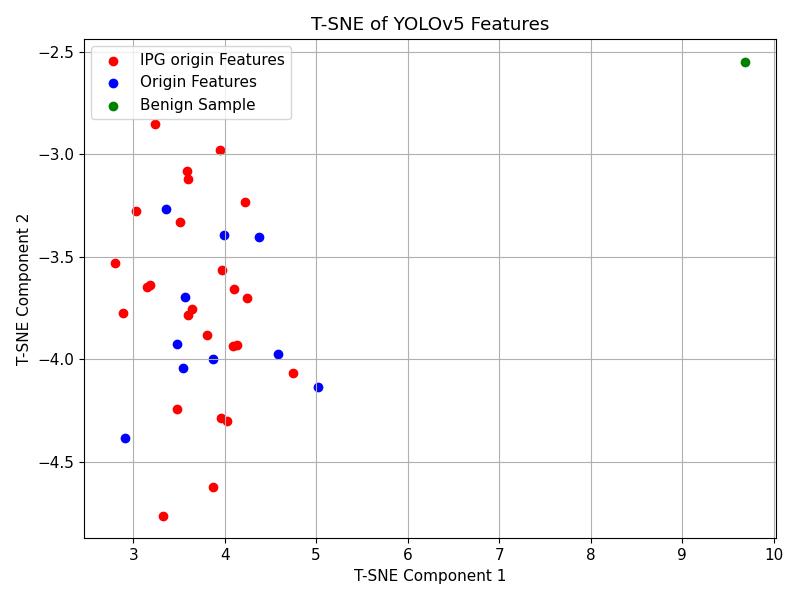}
\caption{Visualization result of Origin Patch and IPG Patch}
\label{fig:result_IPG}
\end{figure}

\subsection{Incremental Patch Generation Result}
Table \ref{table:result_IPG} presents a comparison between the performance of patches generated by the existing method (origin) and those generated using the IPG method. While the existing method required 49.40 hours to generate a single patch, it achieved a higher ASR of 0.669 in comparison to IPG. In contrast, the IPG method generated 25 patches over 112.55 hours, with a relatively lower ASR of 0.600. The existing method yielded an Efficiency of 0.020 patches per hour, whereas IPG demonstrated a markedly higher Efficiency of 0.222 patches per hour, representing 11.1 times the efficiency of the existing method.

Figure \ref{fig:result_IPG} shows the patches generated by the existing method (origin) and those generated using IPG(IPG origin). The distance between two patches and the benign sample is sufficient, indicating the effectiveness of the adversarial patch attack. A comparison of the distribution of features reveals that those generated using IPG are dispersed over a wider range than those of the origin. This indicates that the IPG patches are capable of covering a broader range of vulnerabilities in the target model, Yolov5l6, and exhibit superior generalization capabilities. It can thus be concluded that IPG patches are more suitable for adversarial training than existing adversarial patches.

\section{Ablation Study}
\subsection{Data Amount}
The data amount utilized for the generation of adversarial patches has a considerable impact on ASR. While the use of a larger set of images generally results in enhanced performance for generalized patches, this approach inevitably entails a trade-off in terms of efficiency. To investigate whether effective and generalized adversarial patches can still be generated with smaller datasets, we progressively reduced the sampled dataset size from 550 images by increments of 100 images for patch generation. The performance of the generated patches is shown in Table \ref{table:ablation_DataAmount}.
\begin{table}[t] 
\caption{Ablation Study on Data Amount}
\vspace{3pt}
\label{table:ablation_DataAmount}
\newcolumntype{C}{>{\centering\arraybackslash}X}
\begin{tabularx}{\textwidth}{CCCCC}
\toprule
\textbf{Patch Method}&\textbf{Gen Time(hour)}&\textbf{Patch Amount}&\textbf{Efficiency}&\textbf{ASR}\\
\midrule
IPG\_550      & 112.55         & 25           &  0.222     & 0.600 \\
IPG\_450      &  92.96         & 25           &  0.269     & 0.592 \\
IPG\_350      &  74.65         & 25           &  0.335     & 0.584 \\
IPG\_250      &  52.25         & 25           &  0.478     & 0.579 \\
\bottomrule
\end{tabularx}
\end{table}
\begin{figure}[t]
\centering
\includegraphics[scale=0.50]{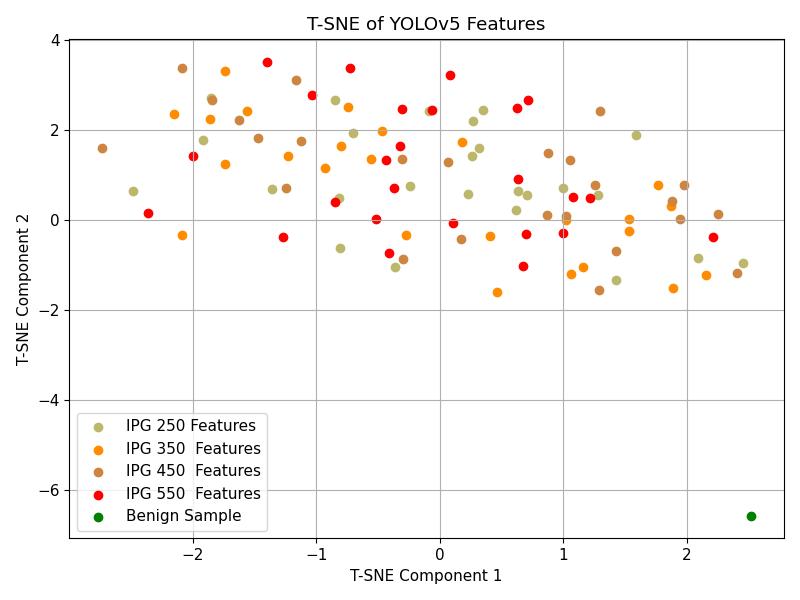}
\caption{Visualization result of IPG's Data Amount}
\label{fig:IPG_DataAmount}
\end{figure}

As the amount of data utilized for patch generation is reduced, the Efficiency of patch generation increases, while the ASR declines. This demonstrates that increasing the amount of data employed for patch generation can also enhance the ASR when employing IPG. As shown in Figure \ref{fig:IPG_DataAmount}, the generalization results based on the amount of data utilized for generation indicate that, in general, all patches are effectively distinguished from benign samples and are widely dispersed. This corroborates the ability of IPG to generate well-generalized adversarial patches with a relatively limited amount of data.

\subsection{Data Augmentation}
\begin{figure}[t]
\centering
\includegraphics[scale=0.6]{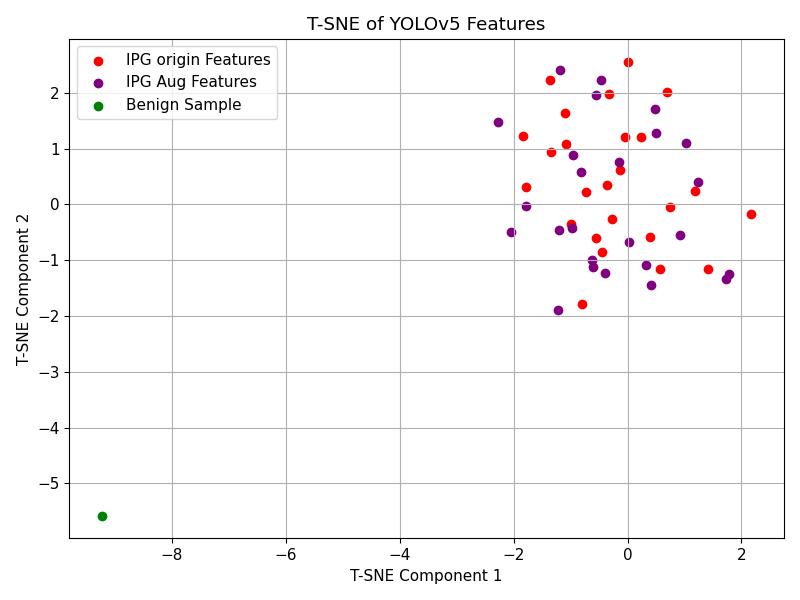}
\caption{Visualization result of IPG\_origin and IPG\_aug} 
\label{fig:IPGorigin_vs_IPGAaug}
\end{figure}

The augmented model's robustness against common occlusions may result in a decrease in the ASR when performing adversarial patch attacks. As shown in Table \ref{table:ablation_DataAug}, the patch Efficiency using IPG did not exhibit a notable change, as the same amount of data was utilized, and the ASR remained comparable. This corroborates the effectiveness of IPG against the augmented model. The visualization of IPG patches generated by the augmented model(IPG\_aug) and the origin model(IPG\_origin) using PCA and T-SNE is presented in Figure \ref{fig:IPGorigin_vs_IPGAaug}. Both types of IPG patches are evenly distributed and are clearly separated from benign samples, indicating that IPG can generate well-generalized patches for the augmented model as well.

\begin{table}[t]
\vspace{3pt}
\caption{Ablation Study on Data Augmentation}
\label{table:ablation_DataAug}
\newcolumntype{C}{>{\centering\arraybackslash}X}
\begin{tabularx}{\textwidth}{CCCCC}
\toprule
\textbf{Patch Method}&\textbf{Gen Time(hour)}&\textbf{Patch Amount}&\textbf{Efficiency}&\textbf{ASR}\\
\midrule
IPG\_origin      & 112.55         & 25           &  0.222     & 0.600 \\
IPG\_aug         & 113.50         & 25           &  0.220     & 0.593 \\
\bottomrule
\end{tabularx}
\end{table}

\section{Adversarial Training using IPG}
To verify whether IPG-generated adversarial patches can actually improve the inherent robustness of AI models and provide a robust knowledge base for sematic interoperability, we conducted adversarial training experiments. The most commonly used method for adversarial training with adversarial patches is to randomly attach adversarial patches to the training dataset to build adversarial data and utilize it in the learning process. In this context, IPG generates many diverse patches simultaneously, which can systematically enhance the semantic coverage of vulnerabilities. Therefore, to check whether IPG is effective for adversarial training, we first built adversarial data by randomly attaching IPG patches to MS COCO 2017 \cite{coco_lin2014} train-set by randomly sampling 2,345 images from the train dataset for 50 patches. To prevent the accuracy of the clean dataset from decreasing during adversarial training, we set the proportion of the adversarial dataset and the clean dataset to 50\% each.

When performing adversarial training, it is important to consider both clean accuracy and attack defense, as well as the semantic interoperability of robustness knowledge. Clean accuracy refers to the performance when no adversarial attack is performed, i.e., after performing adversarial training, the performance of the model (AT) that has undergone adversarial training should not decrease significantly compared to the model (Base) trained on a clean dataset without adversarial patches. Additionally, from the perspective of attack defense and semantic knowledge management, it is crucial to understand and represent the inherent robustness properties of the model clearly.

Therefore, to evaluate the effectiveness of adversarial training in covering a semantically broader vulnerability space, we measured the mAP, mAR, and ASR for the patches (Seen) used for training. The mAP and mAR represent the model’s detection performance; higher values indicate better performance. ASR is the attack success rate of the adversarial patch, and lower values imply higher robustness. Moreover, we assessed the performance when random noise was added alongside adversarial patches and evaluated robustness against general occlusion. Finally, to see if this adversarial training improved the inherent robustness of the model itself, we performed the same adversarial patch attack (Unseen) on the model that had undergone adversarial training and measured the mAP, mAR, and ASR. Such comprehensive evaluations highlight IPG’s potential to support a structured, semantically interoperable knowledge framework, thus significantly contributing to proactive vulnerability recognition and enhanced decision-making within AI security frameworks.

\begin{table}[t] 
\caption{Clean Accuracy of Base Model and IPG Adversarial Training Model}
\vspace{3pt}
\label{table:yolov5l6_AT_performance}
\newcolumntype{C}{>{\centering\arraybackslash}X}
\begin{tabularx}{\textwidth}{CCCCCC}
\toprule
\textbf{Model}&\textbf{mAP50}&\textbf{mAP50-95}&\textbf{Precision}&\textbf{Recall}&\textbf{F1-Score}\\
\midrule
COCO\_Base      & 0.658 & 0.470    &  0.744    & 0.596  & 0.662   \\
COCO\_AT        & 0.655 & 0.468    &  0.709    & 0.608  & 0.653   \\
\bottomrule
\end{tabularx}
\end{table}

The performance for clean accuracy after performing adversarial training with IPG is as shown in Table \ref{table:yolov5l6_AT_performance}. When comparing COCO\_Base without and COCO\_AT in terms of clean accuracy, COCO\_AT with adversarial training shows a relatively lower value. However, the decrease is not large, and the improvement in recall performance can be seen, so overall, it can be said that the performance drop for clean accuracy is not significant. This indicates that ensuring equal distribution of clean data and adversarial data in the training dataset is sufficient to maintain clean accuracy, with the potential for enhancement through the implementation of additional methods.

\begin{table}[t]  
\caption{Adversarial Training Results with IPG}
\vspace{3pt}
\label{table:adversarial_training_results}
\newcolumntype{C}{>{\centering\arraybackslash}X}
\begin{tabularx}{\textwidth}{CCCC|CCC}
\toprule
\multicolumn{1}{C}{\multirow{2}{*}{\textbf{Model}}} & \multicolumn{3}{c|}{\textbf{Random Noise}}                  & \multicolumn{3}{c}{\textbf{Adversarial Patch}}     \\
\multicolumn{1}{c}{} & \multicolumn{1}{c}{mAP50-95($\uparrow$)} & \multicolumn{1}{c}{mAR50-95($\uparrow$)} & \multicolumn{1}{c|}{ASR($\downarrow$)} & \multicolumn{1}{c}{mAP50-95($\uparrow$)} & \multicolumn{1}{c}{mAR50-95($\uparrow$)} & \multicolumn{1}{c}{ASR($\downarrow$)} \\ 
\midrule
COCO\_Base  & 0.348 & 0.408 & 0.525 & 0.263 & 0.310 & 0.621 \\
\midrule
COCO\_AT(Seen) & 0.740 & 0.775 & 0.178 & 0.755 & 0.790 & 0.162 \\
COCO\_AT(Unseen) & 0.739 & 0.775 & 0.178 & 0.300 & 0.360 & 0.588 \\
\bottomrule
\end{tabularx}
\end{table}

The results for the adversarial training and the evaluation of its inherent robustness are shown in Table \ref{table:adversarial_training_results}. The two metrics, COCO\_Base and COCO\_AT(Seen) are evaluated on the exact same IPG adversarial patches, which refers to performance on patches directly utilized in adversarial training. When comparing the two models, the mAP and mAR values for COCO\_AT(Seen) with adversarial training are very high for both random noise and adversarial patches. We can see that the ASR for random noise decreased by about 35\%, from 0.525 to 0.178, while the mAP and mAR improved by about 40\% and 37\%, respectively. For adversarial patch, we can see that the ASR decreased by about 46\% from 0.621 to 0.162, and the mAP and mAR improved significantly by about 49\% and 48\%, respectively. This shows that adversarial training with IPGs can significantly improve the robustness of the model to the IPGs used for training as well as to general occlusions.

COCO\_AT(Unseen) is the result of performing adversarial training with IPG and then performing adversarial patch attack again, which allows us to evaluate the inherent robustness of the model to adversarial training with IPG. Comparing the metrics for random noise between COCO\_Base and COCO\_AT(Unseen), we can see that the mAP and mAR are very high, the same as COCO\_AT(Seen), and the ASR is reduced by about 35\%. This suggests that the adversarial data built using IPG has significantly improved the inherent robustness to general occlusion because it is updated from Gaussian random noise when generating the adversarial patch. Comparing the metrics for the adversarial patch in both models, we can see that ASR is reduced by about 3\%, while mAP and mAR are improved by 3.7\% and 5\%, respectively. Taken together, these results show that adversarial training with IPG significantly improves the inherent robustness of the same adversarial patch attack as well as general occlusion. These results are further beneficial in that these performance metrics were achieved with the least trade-off to clean accuracy.

\section{Conclusion}
In this paper, we propose IPG, a novel and efficient approach to adversarial patch generation. The efficiency of IPG is demonstrated through detailed experiments, which show that it is significantly more efficient than existing methods. It was shown to generate patches up to 11.1 times faster while maintaining competitive ASR. One of the key contributions of this work is the ability of IPG to generate patches that address a wider range of vulnerabilities, as demonstrated by the results of our ablation studies. The generalization of these patches was visualized using PCA and t-SNE, which revealed that IPG patches exhibited greater dispersion compared to patches generated by conventional methods, indicating superior generalization performance. Our results show that IPG is effective not only in standard models but also in augmented models, demonstrating its robustness under different training conditions. Furthermore, by conducting adversarial training with IPG and confirming the inherent improvement in robustness, we show that IPG is suitable for generalized adversarial training on adversarial patches. Additionally, IPG-generated adversarial patches can provide structured knowledge for constructing a robustness model, facilitating advanced interoperability, proactive vulnerability management, and robust decision-making within AI security ecosystems. These results provide a foundation for future work on integrating IPG into adversarial training frameworks, where it can significantly contribute to improving model robustness against adversarial patch attacks and support structured AI defense strategies.

\bibliographystyle{unsrt}
\bibliography{references}

\end{document}